\documentclass{article}

\usepackage{arxiv}

\usepackage{amssymb}
\usepackage{amsmath}
\usepackage[utf8]{inputenc} 
\usepackage[T1]{fontenc}    
\usepackage{hyperref}       
\usepackage{url}            
\usepackage{booktabs}       
\usepackage{amsfonts}       
\usepackage{nicefrac}       
\usepackage{microtype}      
\usepackage{xcolor}         
\usepackage{algorithm}
\usepackage{algpseudocode}
\usepackage{graphicx}
\usepackage{multirow}
\usepackage[flushleft]{threeparttable}
\usepackage{float}
\usepackage{caption}

\title{Being Strong Progressively! Enhancing Knowledge Distillation of Large Language Models through a Curriculum Learning Framework}

\author{
 Lingyuan Liu \\
  City University of Hong Kong \\
  \texttt{ly.liu@my.cityu.edu.hk} \\
   \And
 Mengxiang Zhang\footnote{$*$} \\
  The University of Hong Kong \\
  \texttt{mxzhang6@connect.hku.hk} \\
}
\begin{document}
\maketitle
\begin{abstract}
Knowledge Distillation (KD) compresses large language models (LLMs) by transferring the teacher model’s capabilities to a smaller student model, reducing inference cost and memory usage while maintaining performance. However, existing KD methods for LLMs often fail to prevent significant shifts in the student model’s distribution during training, leading to issues such as catastrophic forgetting, mode collapse, and training-inference mismatch. To address these challenges, we propose a novel, plug-in curriculum learning framework inspired by the strength training principle of "progressive overload" (POCL), which can be seamlessly integrated into existing white-box KD approaches with minimal computational overhead. 
The framework comprises two core components: (1) a difficulty measurer that ranks and partitions training samples from easy to hard, and (2) a training scheduler that incrementally introduces these subsets into the distillation process at fixed intervals while applying loss functions with progressively rising temperatures. By starting with the easiest samples and progressively increasing the difficulty, the approach enhances both the stability and efficiency of learning. Extensive experiments in instruction-following settings demonstrate that POCL consistently improves the performance of distilled student models across various white-box KD methods and model families. Our findings highlight the effectiveness of sorted training samples in KD for LLMs. More generally, our work demonstrates how to structure training data within the KD process to enhance the stability and performance of distilled LLMs.
\footnotetext{Corresponding author.}
\footnote{The code for our method is publicly available at \href{https://github.com/liuliuyuan6/POCL}{https://github.com/liuliuyuan6/POCL}.}
\end{abstract}

\begin{figure}[!ht]
   \begin{center}
     \includegraphics[scale = 0.065]{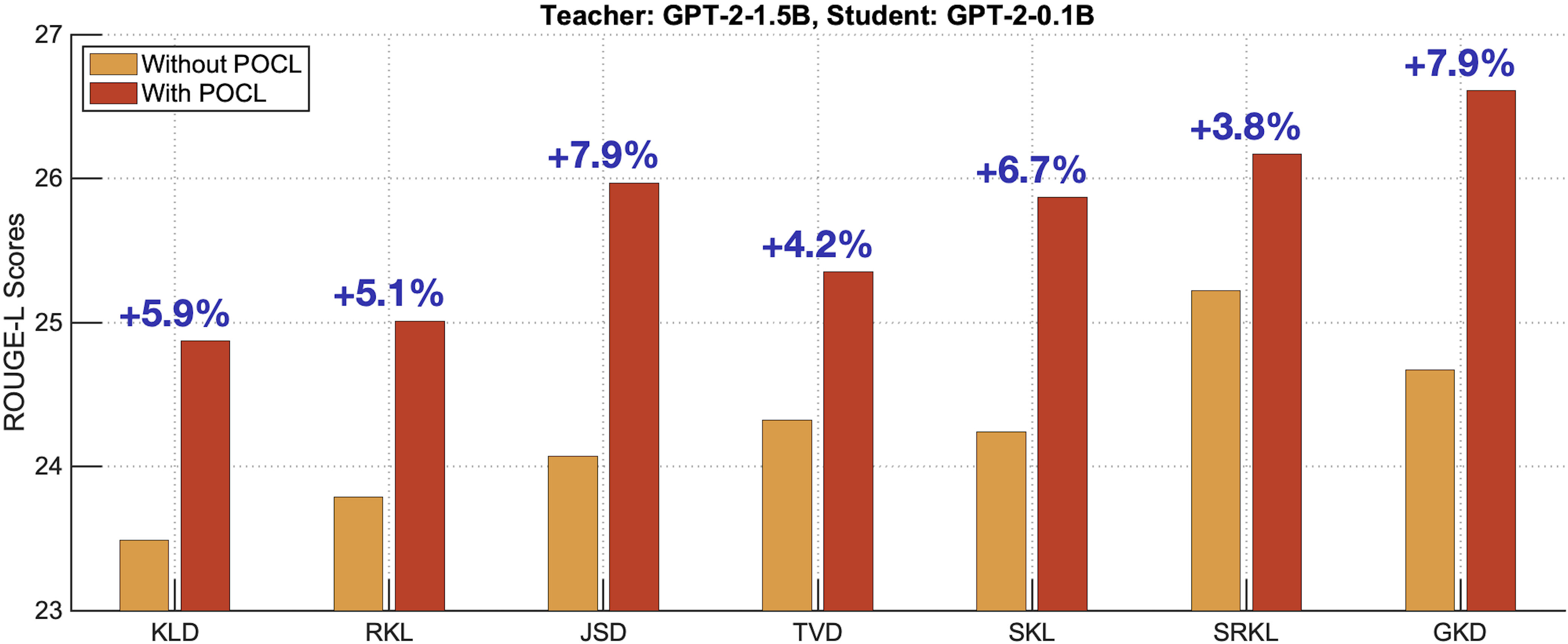}
   \caption{\small Effectiveness of the POCL framework across KD methods on DollyEval.
This figure compares off-policy (KLD \cite{hinton2015distilling}, RKL, JSD, TVD \cite{wen2023f}, SKL and SRKL \cite{ko2024distillm}) and on-policy (GKD \cite{agarwal2024policy}) KD methods with and without POCL, evaluated using ROUGE-L scores. The setup uses GPT-2-1.5B $\rightarrow$ GPT-2-0.1B. POCL improves all base white-box KD methods.}\label{fig:main_result}
   \end{center}
   \end{figure}

\section{Introduction}\label{sec:01}

Large language models (LLMs) have significantly advanced in text generation, language understanding, and inference capabilities through increases in parameter count and the use of high-quality training data \cite{ouyang2022training}. These models are now widely applied across diverse domains. However, their deployment is challenging due to high computational and memory demands during inference, requiring powerful hardware such as GPUs and substantial storage—hindering practical implementation, particularly in edge computing environments \cite{aryan2023costly}. Consequently, there is a growing need for compact, efficiently deployable LLMs that maintain competitive performance in key areas like text generation \cite{li2024pre} and tool learning tasks \cite{gao2024confucius}.

Knowledge distillation (KD) \cite{hinton2015distilling}, a widely used model compression technique in deep learning, transfers knowledge from a high-capacity teacher model to a compact student model. With the advancement of LLMs, KD has been increasingly applied to develop efficient small language models (LMs) that retain strong performance, as seen in recent works such as Llama 3.2 \cite{meta2024llama32}, Gemma-2 \cite{team2024gemma}, and Deepseek-R1 Distilled Models \cite{deepseekai2025deepseekr1incentivizingreasoningcapability}.

In the era of LLMs, KD methods are broadly categorized into black-box KD - being akin to data augmentation \cite{xu2024survey} and white-box KD \cite{yang2024survey}. Black-box KD, which transfers knowledge solely through teacher model predictions \cite{kim2016sequence}, has been widely adopted due to the prevalence of proprietary LLMs such as GPT-4o \cite{hurst2024gpt}, Claude 3.5 \cite{claude35_sonnet_news}, and Gemini 1.5 \cite{team2024gemini}, which restrict access to internal mechanisms. This approach typically leverages techniques like In-Context Learning \cite{wang2023learning}, Chain-of-Thought \cite{hsieh2023distilling}, and Instruction Following \cite{peng2023instruction}. However, recent advances in open-source LLMs—such as DeepSeek-v3 \cite{liu2024deepseek} and Qwen 2.5 \cite{yang2024qwen2}—have significantly narrowed the performance gap with proprietary models, now at just 1.7\% on the Chatbot Arena Leaderboard \cite{AIIndex2025}. As a result, there is growing interest in white-box KD, which offers superior student model performance and greater control over the distillation process.

\begin{figure}[!ht]
   \begin{center}
     \includegraphics[scale = 0.9]{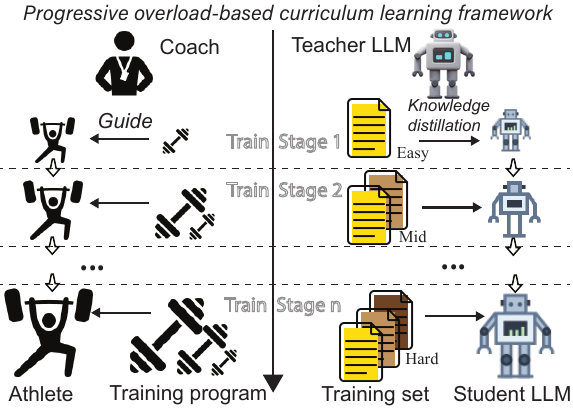}
   \caption{\small Progressive Overload-Based Curriculum Learning (POCL) Framework for KD. Drawing on principles from strength training, POCL models the teacher-student KD process as a coach-athlete training regimen. The training set’s size and difficulty correspond to training volume and intensity, respectively. The student model begins with small, simple data subsets and progressively advances through increasingly complex and larger training stages to enhance learning efficiency.}\label{fig:POCL}
   \end{center}
   \end{figure}

White-box KD for LLMs leverages internal teacher model data during training to enhance student model performance. Recent efforts focus on optimizing loss functions—such as Kullback-Leibler divergence (KLD) and its variants \cite{gu2023minillm, hinton2015distilling}—to balance mode-seeking and mode-averaging behaviors, improving KD effectiveness. Additionally, studies explore data curation strategies, including offline and online data selection \cite{lin2020autoregressive}, to mitigate training-inference mismatches and enhance training efficiency. Algorithmic improvements have also been proposed to stabilize the training process. However, most approaches overlook the influence of training data quality and characteristics on KD performance. Suboptimal combinations of loss formulations and data strategies may limit student model improvement, particularly due to capacity gaps between teacher and student models, which can cause undesirable distributional shifts. Few works, such as \cite{xu2024speculative}, explicitly address this issue.

Curriculum learning (CL), a training strategy that progresses from simpler to more complex data, has been widely applied—particularly in computer vision—to enhance model generalization and convergence \cite{wang2021survey}. As a versatile plug-in method, it has shown effectiveness across various machine learning paradigms. However, its application to KD for LLMs remains underexplored. Our theoretical analysis suggests that CL can effectively address training-inference mismatches arising from capacity gaps between teacher and student models, as well as complement different loss formulations and data curation strategies in LLM distillation. This insight motivates the design of a curriculum learning framework to enhance KD performance in LLMs.

In this paper, we propose a CL framework to enhance KD for LLMs, inspired by the "progressive overload" principle in strength training, where training intensity and volume are gradually increased to build capacity \cite{zatsiorsky2020science}. In our analogy, the teacher model acts as a coach and the student model as an athlete. The training set corresponds to the workout plan, with its size and difficulty representing training volume and intensity, respectively. The KD process is structured as a progressive training regimen: starting with small, simple datasets and incrementally increasing complexity and scale across training stages (See Fig. \ref{fig:POCL}). This approach aligns with CL’s natural ability to mitigate training-inference mismatches caused by capacity gaps between teacher and student models. As a plug-and-play module for white-box KD methods, our framework achieves notable performance improvements with minimal additional computational cost.

\begin{itemize}
    \item \textbf{Progressive Overload-Based Curriculum Learning (POCL) Framework:} To address key challenges in knowledge distillation (KD) for large language models (LLMs)—including training-inference mismatches and limited generalization—we propose POCL, a plug-and-play curriculum learning framework inspired by the "progressive overload" principle in strength training. The framework leverages the student model to assess and rank sample difficulty using reciprocal rank fusion, partitioning the dataset into easy-to-hard subsets. A training scheduler—referred to as Baby Step—iteratively expands the training set, starting from the simplest samples and gradually incorporating more difficult ones after fixed intervals or convergence, until the full dataset is utilized.
    
    \item \textbf{Advanced Performance and Efficiency:} POCL enhances student LLM performance across diverse generative tasks, such as instruction following and text summarization, under various white-box KD settings (See Fig. \ref{fig:main_result}), while introducing minimal additional computational overhead .
\end{itemize}

\section{Background and Preliminary Study}\label{sec:02}

\subsection{Current Framework for White-Box KD}\label{sub:0201}
We define the white-box KD framework for auto-regressive LMs. The student model minimizes:
\begin{equation}\label{eq.01}
    L_{s} = \alpha \cdot L_{ce} + (1-\alpha) \cdot L_{kd},
\end{equation}
where $L_{ce}$ is the cross-entropy loss against ground truth $y$, and $L_{kd}$ measures divergence between teacher $p(y|x)$ and student $q_\theta(y|x)$ distributions scaled by temperature $\tau$. Current methods focus on $D(\cdot||\cdot)$ choices (e.g., KLD \cite{hinton2015distilling}, JSD \cite{agarwal2024policy}) and data curation strategies (e.g., ground truth responses, teacher/student-generated outputs (TGOs vs SGOs) \cite{agarwal2024policy, ko2024distillm}). Despite these efforts, inherent limitations persist. The detailed description of white-box KD is shown in appendix \ref{app:00}.

\subsection{Pitfalls of Existing Distillation}\label{sub:0202}

\textbf{Catastrophic Forgetting and Mode Collapse in Distilled Student LMs.}
A key limitation of current white-box KD frameworks is the risk of catastrophic forgetting or mode collapse in distilled student models. Due to their smaller parameter capacity, student models often struggle to retain previously acquired capabilities during distillation, especially when there is a large capacity gap between teacher and student models \cite{zhang2024dual}. This issue arises because the student model’s distribution diverges from the teacher’s, leading to parameter overwriting, performance drops on prior tasks, or generation of repetitive and biased outputs.
Moreover, KD loss design faces a critical trade-off: losses like reverse KLD, which emphasize sharpness, can prevent over-smoothing but may worsen catastrophic forgetting and mode collapse \cite{li2024revisiting}. Conversely, flat-loss formulations like standard KLD stabilize existing knowledge but yield suboptimal distillation performance. These challenges highlight the difficulty of maintaining a stable student model distribution during training, underscoring a fundamental limitation in current white-box KD approaches.

\textbf{Training-Inference Mismatch in Distilled Student LMs.}
A key challenge in KD for LLMs is the training-inference mismatch: student models are typically trained on fixed datasets, which may not reflect the diverse inputs encountered during inference \cite{ko2024distillm}. To address this, recent methods incorporate SGOs into training, aiming to bridge the gap between training and real-world conditions. However, SGO usage presents a trade-off. While moderate SGO integration improves generalization, excessive reliance can lead to performance degradation due to distribution mismatches between teacher model training data and noisy SGOs, potentially misguiding the distillation process. Moreover, generating SGOs is computationally expensive, often accounting for up to 80\% of total training time \cite{ko2024distillm}. Conversely, insufficient SGO use limits their benefits, resulting in suboptimal KD performance. Thus, an effective white-box KD framework must carefully balance SGO usage to maximize gains while minimizing noise-induced degradation.

\subsection{Potentials of CL for White-Box KD}\label{sub:0203}
Our theoretical analysis suggests that integrating CL into white-box KD can effectively address two major challenges: catastrophic forgetting / mode collapse and training-inference mismatch.

\textbf{CL guides the KD training process.}
CL improves KD by progressively exposing the student model to increasingly complex data, preventing abrupt distribution shifts between teacher and student models \cite{wang2021survey}. By starting with easy samples—where student and teacher distributions align most closely—and gradually introducing harder ones, CL ensures smoother adaptation and mitigates catastrophic forgetting or mode collapse. Additionally, CL accelerates convergence by regularizing training toward favorable regions in parameter space, which benefits on-policy KD methods that use computationally expensive SGOs.

\textbf{CL denoises the training set.}
CL inherently prioritizes high-confidence, low-noise samples early in training, improving robustness and generalization. In the context of KD, this reduces reliance on low-quality SGOs, which are more prevalent when student models generate responses to difficult prompts. A difficulty measurer within CL can further rank and filter SGO quality, enhancing KD performance while limiting noise-induced degradation.

\textbf{CL is flexible and plug-and-play.}
As CL only involves ranking sample difficulty, it does not interfere with core KD components such as loss function design or data curation strategies. This makes it a versatile plug-in compatible with diverse white-box KD approaches, offering consistent improvements at minimal computational overhead.

\section{Methodology}\label{sec:03}
We focus on conditional text generation, where LLMs generate responses $y$ conditioned on prompts $x$. In white-box KD, the student model is trained to minimize a combined objective $L_{s}$ (see Eq.~\ref{eq.01}), consisting of a cross-entropy loss against ground truth and a KD loss measuring divergence from a fixed teacher model $p(y|x)$. The KD framework allows for various loss formulations (e.g., KLD, RKL, JSD) and data curation strategies (e.g., ground-truth labels, teacher-generated outputs, or student-generated outputs).

To enhance this framework, we propose POCL, a plug-and-play CL method inspired by the "progressive overload" principle in strength training \cite{zatsiorsky2020science}. POCL integrates seamlessly into existing white-box KD approaches and consists of two core components:

1. \textbf{Difficulty Measurer Module:} Ranks training samples by difficulty using the student model’s confidence scores via reciprocal rank fusion, partitioning the dataset into easy-to-hard subsets.

2. \textbf{Training Scheduler Module:} Gradually introduces harder samples during training—starting with the easiest subset and incrementally merging in more difficult ones until the full dataset is used. Loss functions with progressively increasing temperatures are applied across successive training stages.

This staged exposure ensures stable learning and mitigates distribution shifts between teacher and student models. Additionally, key KD parameters are adaptively adjusted during training to further boost performance. Algorithmic details are presented in Algorithm~\ref{alg:cap}, with subsections detailing the difficulty measurer (Section~\ref{subsec:0301}), training scheduler (Section~\ref{subsec:0302}), and adaptive parameter control (Section~\ref{subsec:0303}).

\begin{algorithm}\small
\caption{POCL: progressive overload-based curriculum learning for white-box KD algorithm}\label{alg:cap}
\begin{algorithmic}
\State \textbf{Input:} $D$: training dataset, $lm_{s}$: student LM, $lm_{t}$: teacher LM; 
\State \textbf{Output:} $lm_{s}^{*}:$ the distilled student LM.
\State $\{d_{1}, d_{2}, ..., d_{n} \}$ = $D'$ = sort($D$) using Eq. \ref{eq:04};
\State $D_{train} = \emptyset$, $\tau = 1$, $\alpha = 1$;
\State $lm_{s}^{*}$ = $lm_{s}$;
\For{$i = 1, ..., n$}
\State $D_{train} = D_{train} \bigcup d_{i}$;
\State $\tau = \tau(i)$, $\alpha = \alpha(i)$ using Eq. \ref{eq:05} and Eq. \ref{eq:06};
    \While{not converged for $p$ epochs}
    \State $lm_{s}^{*}$ = KD train($lm_{s}^{*}$, $lm_{t}$, $D_{train}$, $\tau$, $\alpha$);
    \EndWhile
\EndFor
\end{algorithmic}
\end{algorithm}

\subsection{Difficulty Measurer}\label{subsec:0301}
In the Difficulty Measurer module, we assess the difficulty of each training sample by combining two metrics: (1) the Rouge-L score \cite{lin2004rouge} between student-generated outputs $y_s$ and ground-truth responses $y$, and (2) the cross-entropy loss $L_{ce}$ (see Eq.~\ref{eq:ce}) computed over the student model’s token-level predictions. Higher Rouge-L scores and lower cross-entropy values indicate easier samples.

We generate two ranked lists—$I_{rl}$ based on Rouge-L and $I_{ce}$ based on cross-entropy—and combine them using reciprocal rank fusion \cite{cormack2009reciprocal}, widely used for re-rank in information retrieval, to produce a final ranking $I_{dm}$. This method computes a fused score for each sample as:

\begin{equation}\small\label{eq:04}
FR_{score} =  \sum_{i}^{n} \frac{1}{k+r_{i}},
\end{equation}
where $r_i$ is the rank in list $i$, and $k = 60$ is a constant to dampen outlier rankings \cite{cormack2009reciprocal}. A higher $FR_{{score}}$ indicates an easier sample.

Using $I_{dm}$, we partition the training set $D$ into $n$ subsets $\{d_1, \dots, d_n\}$ from easiest to hardest. Each subset contains approximately $\lfloor N/n \rfloor$ or $\lfloor N/n \rfloor + 1$ samples, where $N$ is the total number of samples. We empirically find that setting $n = 4$ is a generally good value and do not tune it on a dataset basis.

\subsection{Training Scheduler}\label{subsec:0302}
In the Training Scheduler module, we implement a discrete scheduler called Baby Step \cite{bengio2009curriculum} that controls the progression of training subsets used during KD. Given $n$ pre-divided training subsets $\{d_1, \dots, d_n\}$ ordered from easiest to hardest (see Section~\ref{subsec:0301}), the KD process is divided into $n$ stages.

At each stage $i$, the student model is trained on the union of subsets $\bigcup_{j=1}^{i} d_j$. Training begins with the easiest subset $d_1$, and after a fixed number of epochs or convergence, the next subset $d_{i+1}$ is incrementally added. This gradual expansion continues until all subsets are incorporated and training concludes. See Algorithm~\ref{alg:cap} for full details.

Loss functions with increasing temperatures are applied across successive stages. The distillation temperature $\tau$ (see Eq.~\ref{eq:tem_q} and \ref{eq:tem_p}), which governs the smoothness of teacher and student output distributions in KD losses (see Eq.~\ref{eq:kd}), starts at a low value and linearly rises over stages:

\begin{equation}\small\label{eq:05}
    \tau_i = \tau_{0} + (\tau_{n} - \tau_{0}) \cdot \frac{i-1}{n-1},
\end{equation}
where $\tau_{0} = 1$ and $\tau_{n} = 2$ are empirically fixed values. This schedule enables the student to initially focus on confident predictions and progressively learn from softer, more global teacher signals.

\subsection{Adaptive parameters for KD methods}\label{subsec:0303}

To further enhance the performance within the POCL framework, we introduce adaptive control of the supervised fine-tuning (SFT) ratio $\alpha$, which controls the relative weight of cross-entropy and KD losses in the total objective, follows an inverse trend: starting high and decreasing linearly:

\begin{equation}\small\label{eq:06}
    \alpha_i = \alpha_{0} - (\alpha_{0} - \alpha_{n}) \cdot \frac{i-1}{n-1},
\end{equation}

We set $\alpha_{0} = 0.3$ and $\alpha_{n} = 0$ for off-policy KD methods, prioritizing ground-truth learning early and teacher knowledge later, inspired by least-to-most prompting strategies in cognitive education \cite{libby2008comparison}. For on-policy KD, we fix $\alpha = 0$ throughout training, eliminating the cross-entropy component. These adaptive schedules improve stability and performance.

\section{Experiments}\label{sec:04}

\subsection{Experimental Setup}\label{subsec:04-01}

We evaluate the effectiveness of our POCL framework through instruction-following tasks, where models are trained to generate task-compliant responses based on given instructions. The experimental pipeline involves first fine-tuning a large LLM on a training set $D$ of instruction-response pairs to serve as a teacher model, and then KD into smaller student models using various KD methods—with and without POCL—followed by performance comparison \cite{gu2023minillm}.

\textbf{Base Models.} We select two LLM families at different scales as student models: GPT-2 (120M) and OPT (350B), with their larger counterparts (GPT-2 1.5B and OPT 2.7B) serving as teacher models, respectively. Further model details are provided in Appendix~\ref{app:0101}.

\textbf{Training.} We use the \texttt{databricks-dolly-15K} dataset \cite{DatabricksBlog2023DollyV2}, consisting of 15K human-written instruction-response pairs. After filtering out samples exceeding context length limits, we split the dataset into 12.5K training, 1K validation, and 0.5K test samples. Hyperparameters are selected based on Rouge-L scores on the validation set \cite{lin2004rouge}. For on-policy KD, we use a mix of 50\% SGOs and ground truth responses. Using Eq.~(\ref{eq:04}), the 12.5K training samples are ranked and divided into $n = 4$ difficulty-based subsets. All models are trained for the same number of total iterations for fair comparison. On-policy methods use $\alpha = 0$, while off-policy methods apply adaptive balancing with $\alpha_{\text{0}} = 0.3$ and $\alpha_{\text{n}} = 0$. See Appendix~\ref{app:0102} for full training settings.

\textbf{Evaluation.} Distilled models are evaluated on five instruction-following datasets: DollyEval, SelfInst, VicunaEval, S-NI, and UnNI \cite{DatabricksBlog2023DollyV2, wang2022self, chiang2023vicuna, wang2022benchmarking, honovich2022unnatural}. Rouge-L scores are used as the primary metric due to their strong correlation with human judgment \cite{agarwal2024policy}. 
Responses are generated with temperature = 1, and we report average scores over five generations per prompt using five random seeds. Evaluation details are included in Appendix~\ref{app:0103}.

\textbf{Baselines.} We compare against three baselines and the detailed descriptions are shown in Appendix~\ref{app:0104}:
\begin{itemize}
    \item \textbf{SFT}: Direct fine-tuning on ground-truth responses.
    \item \textbf{SeqKD} \cite{lin2020autoregressive}: Student training using teacher-generated outputs (TGOs).
    \item \textbf{White-box KD} \cite{hinton2015distilling}: Token-level distillation from teacher distributions. We examine both loss formulations and data curation strategies: GKD(on-policy) \cite{agarwal2024policy}, KLD \cite{hinton2015distilling},  reverse KLD \cite{gu2023minillm}, JSD, total variation distance \cite{wen2023f}, skew KLD\cite{ko2024distillm}, and skew reverse KLD\cite{ko2024distillm}.
\end{itemize}

\begin{table}\centering\scriptsize
\caption{ Evaluation of the POCL framework. Rouge-L scores on several benchmarks with GPT2-120M as the student. }\label{tab:01}
\begin{threeparttable}
\begin{tabular}{ccl|cccccc} \toprule
Model & \# Params & KD Methods & \textbf{DollyEval} & \textbf{SelfInst} & \textbf{VicunaEval} & \textbf{S-NI} & \textbf{UnNI} & \textbf{Avg.} \\ \midrule
                        &                     1.5B & Teacher                      &27.19&	14.04&	16.47 &27.66 &31.86 &23.44\\ \cmidrule(r){2-9} 
                        &                          & SFT                          &23.33&	10.01&	14.72 &16.38 &19.57 &16.80\\
                        &                          & SeqKD                        &23.72&	11.23&	14.31 &16.48 &19.81 &17.11\\ 
                        &                          & GKD (on-policy)                          &24.67&	11.48&	15.66 &23.81 &25.26 &20.17\\
                        &                          & ~~~~~~\tiny{+POCL}                      &\textbf{26.61}&	12.62&	16.73 &27.02 &29.61 &22.51 (2.33 $\uparrow$)\\
                        &                          & KLD                          &23.49&	10.33&	14.96 &19.71 &22.01 &18.10\\
                        &                          & ~~~~~~\tiny{+POCL}                    &24.87&	11.56&	16.13 &21.59 &24.34 &19.70 (1.60 $\uparrow$)\\
                        &                          & RKL                          &23.79&	12.13&	14.94 &23.81 &22.52 &19.49\\
                        &                          & ~~~~~~\tiny{+POCL}                      &25.01&	12.76&	16.01 &25.63 &25.42 &20.97 (1.47 $\uparrow$)\\
                        &                          & JSD                          &24.07&	11.38&	15.87 &22.84 &23.06 &19.39\\
                        &                          & ~~~~~~\tiny{+POCL}                    &25.97&	11.74&	~~ \textbf{16.77}$^*$  &26.61 &25.21 &21.26 (1.87 $\uparrow$)\\
                        &                          & TVD                          &24.32&	11.09&	15.51 &25.93 &26.55 &20.67 \\
                        &                          & ~~~~~~\tiny{+POCL}                    &25.35&	13.19&	16.17 & ~~\textbf{28.98}$^*$ &30.09 &22.76 (2.08 $\uparrow$)\\
                        &                          & SKL                          &24.24&	12.27&	15.71 &23.33 &24.02 &19.91\\
                        &                          & ~~~~~~\tiny{+POCL}                     &25.87&	13.08&	16.45 & ~~28.35$^*$  &28.79 &22.51 (2.59 $\uparrow$)\\
                        &                          & SRKL                         &25.22&	12.86&	15.18 &25.51 &28.43 &21.44\\
\multirow{-16}{*}{GPT-2} & \multirow{-15}{*}{0.1B}   & ~~~~~~\tiny{+POCL}                    &26.17&	\textbf{13.28}&	~~16.66$^*$ &~~28.49$^*$ &\textbf{30.12} &\textbf{22.94} (1.51 $\uparrow$)\\ \bottomrule
\end{tabular}
    \begin{tablenotes}\scriptsize
      \item[] Note: GKD \cite{hinton2015distilling} uses JSD as the KD loss and employs an on-policy data curation strategy, whereas JSD refers to the use of JSD as the KD loss with an off-policy data curation approach. We list the mean values among 5 random seeds. The best scores among student models are \textbf{bold}, and the scores where the student model outperforms the teacher are marked with $*$. RKL, TVD, SKL, and SRKL denote reverse KLD, total variation distance, forward and reverse skew KLD, respectively. \textbf{Avg.} denotes the average ROUGE-L score across five evaluation datasets. Results with OPT-350M as the student can be found in Tab. \ref{tab:a01} (Appendix~\ref{app:02}) .
    \end{tablenotes}
\end{threeparttable}
\end{table}

\subsection{Results}\label{subec:0402}
We present Rouge-L evaluation results in Tables~\ref{tab:01} and \ref{tab:a01} (Appendix~\ref{app:02}). Comparing baselines with and without the POCL framework, we find that models distilled using POCL consistently outperform near all baselines across both base models (GPT-2 and OPT), multiple KD methods-GKD (on policy KD) and KLD, JSD, TVD, SKL, SRKL (off policy KD), and all evaluation datasets. This demonstrates POCL’s strong generalization and compatibility with diverse white-box KD approaches, whether based on loss function design or data curation strategy. Moreover, POCL yields particularly notable improvements on datasets beyond DollyEval, indicating enhanced cross-domain generalization. 

\begin{figure}[!ht]
   \begin{center}
     \includegraphics[scale = 0.35]{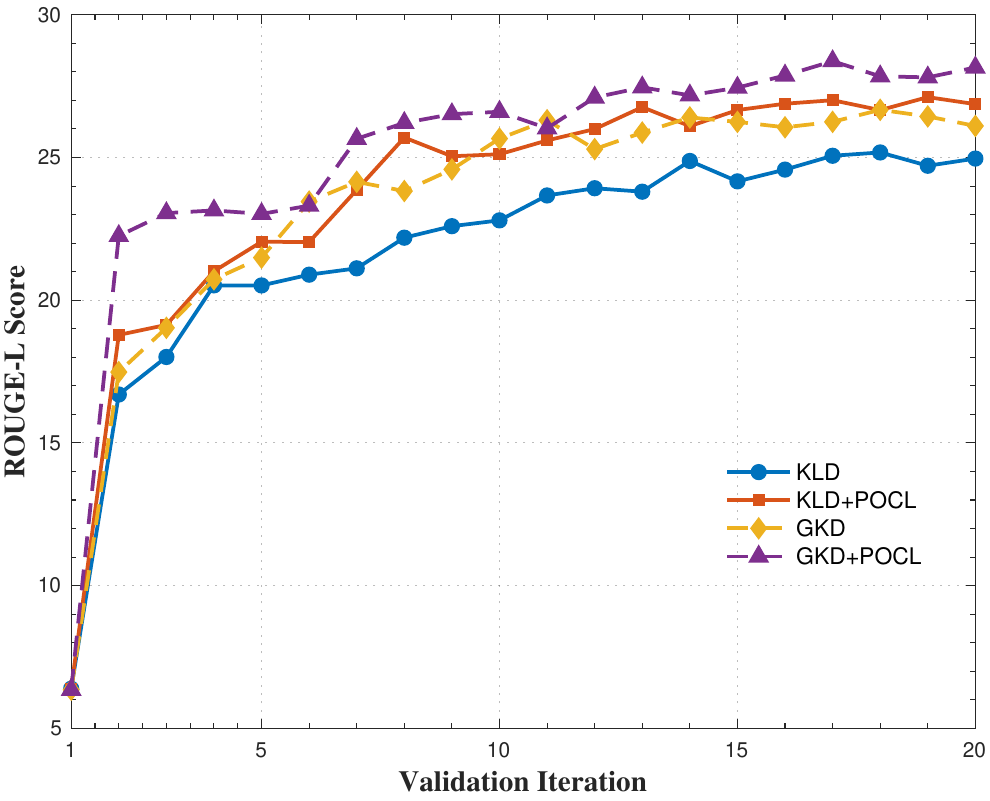}
   \caption{\small ROUGE-L scores for KD vs. KD+POCL on dolly validation set.}\label{fig:RL}
   \end{center}
   \end{figure}

As shown in Fig.~\ref{fig:RL}, both off-policy (KLD) \cite{hinton2015distilling} and on-policy (GKD) \cite{agarwal2024policy} KD methods achieve significantly higher validation Rouge-L scores throughout training when combined with POCL. These results highlight POCL's ability to improve performance stability, underscoring its effectiveness as a plug-and-play enhancement for knowledge distillation of LLMs. 

\section{Analysis and Discussion}\label{sec:05}
To better understand the effectiveness of the POCL framework, we conduct ablation studies using the GPT-2 and OPT model families on the Dolly dataset. We analyze the impact of each component, starting with the Difficulty Measurer module.

\subsection{Effect of Difficulty Measurer}\label{subsec:0501}

The Difficulty Measurer ranks training samples by difficulty using a fusion approach (Eq.~\ref{eq:04}), combining Rouge-L scores ($I_{rl}$) and cross-entropy loss values ($I_{ce}$) via reciprocal rank fusion \cite{cormack2009reciprocal}. In this ablation study, we compare this fusion method against using only Rouge-L or only cross-entropy for ranking.
We apply these ranking strategies to KLD-based knowledge distillation and compare performance across in-domain and out-of-domain datasets (see Tables~\ref{tab:04} and \ref{tab:a04} in Appendix~\ref{app:02}). Results show that the fusion-based ranking consistently outperforms single-metric rankings. Moreover, all variants of KLD with POCL outperform standard KLD without POCL, demonstrating the benefit of curriculum-based sample ordering.
These results highlight the importance of accurate difficulty estimation and validate the effectiveness of reciprocal rank fusion in improving ranking reliability within the POCL framework.

\begin{table}\centering\scriptsize
\begin{threeparttable}
\caption{Performance comparison of KLD and KLD+POCL using different ranking approaches on the DollyEval dataset.}\label{tab:04}
\begin{tabular}{l|cccc} \toprule
Model  & \multicolumn{4}{c}{KD Methods} \\
\# Params (T/S)  &  KLD   & KLD+POCL (R-L) & KLD+POCL (Loss) &  KLD+POCL (Fusion)   \\\midrule
GPT-2 (1.5B/0.1B) & 23.49 &	24.56 &	24.27 &	24.87\\
OPT (2.7B/0.3B)   & 23.09 &	25.23 &	24.22 &	25.45\\ \bottomrule
\end{tabular}
    \begin{tablenotes}\scriptsize
      \item[] Note: 'Params (T/S)' denotes the number of parameters in the teacher and student models. 'R-L' indicates that training samples are sorted by ROUGE-L score \cite{lin2004rouge}, 'Loss' by cross-entropy loss, and 'Fusion' by the Fusion rank (see Eq. \ref{eq:04}). Detailed results are listed in Table~\ref{tab:a04}. 
    \end{tablenotes}
\end{threeparttable}
\end{table}

\subsection{Effect of Training Scheduler}\label{subsec:0502}

\textbf{Effect of Adaptive Parameters.}  
We evaluate the impact of adaptive parameter control— distillation temperature and SFT ratio—in the Training Scheduler module of POCL, focusing on both off-policy (KLD, JSD) and on-policy (GKD) KD methods. Results across five evaluation datasets (see Tables~\ref{tab:03} and \ref{tab:a03} in Appendix~\ref{app:02}) indicate that the distillation temperature plays a critical role in the POCL framework. Specifically, the pure POCL framework—without adaptive parameters—or its variant using only the SFT ratio, underperforms compared to baselines without POCL. In contrast, the integration of rising distillation temperature achieves superior performance over all baselines. Notably, temperature has a more pronounced effect than the SFT ratio. For on-policy KD, we observe that adaptive SFT ratios provide no benefit, as student learning should prioritize teacher guidance over ground-truth labels. These findings underscore the importance of adaptive parameter scheduling, particularly for temperature, in achieving stable and effective KD within the POCL framework.

\begin{figure}[!ht]
   \begin{center}
     \includegraphics[scale = 1.1]{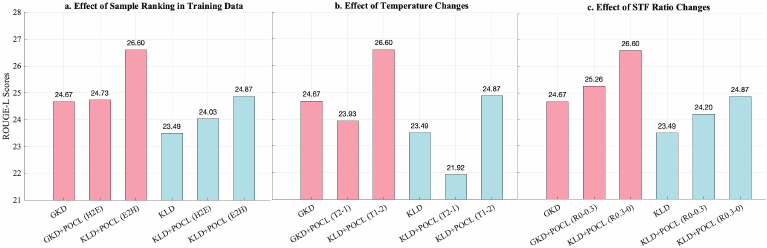}
   \caption{\small Ablation study of key POCL components—training sample ranking, distillation temperature, and STF ratio—on the performance of the distilled student model across different KD methods. In the figures: (a) 'E2H' ('easy-to-hard') and 'H2E' ('hard-to-easy') indicate the order of training samples; (b) 'T2-1' and 'T1-2' denote initial and final temperature values of 2 and 1, respectively; (c) 'R0-0.3' and 'R0.3-0' represent initial and final SFT ratios of 0 and 0.3, respectively.}\label{fig:ETS}
   \end{center}
   \end{figure}

\textbf{Effect of Training Sample Ranking.}  
We evaluate how different training sample ordering strategies in the Training Scheduler module affect KD performance. As shown in Figure~\ref{fig:ETS}(a) and Table~\ref{tab:a05} (Appendix~\ref{app:02}), POCL consistently outperforms baselines regardless of whether samples are ordered from easy-to-hard or hard-to-easy, highlighting the general benefit of structured data exposure during training. However, easy-to-hard scheduling achieves superior results, suggesting that early exposure to complex samples may hinder later learning of simpler patterns.

\begin{table}\centering\scriptsize
\begin{threeparttable}
\caption{Performance comparison of KLD+POCL with and without adaptive parameters on the five evaluation datasets.}\label{tab:03}
\begin{tabular}{l|ccccc} \toprule
 KD Methods  & \textbf{DollyEval} & \textbf{SelfInst} & \textbf{VicunaEval} & \textbf{S-NI} & \textbf{UnNI} \\ \midrule
 KLD                           & 23.49& 10.33&  14.96&  19.70&  22.01 \\
~~+POCL (w/o temp. \& ratio)   & 23.21&	10.42&	14.79&	18.97&	21.85 \\
~~+POCL (w/o temp.)            & 22.04&	 8.70&	14.93&	19.63&	21.77 \\
~~+POCL (w/o ratio)            & 24.33&	11.47&	15.76&	20.98&	23.87 \\
~~+POCL                        & 24.87&	11.56&	16.13&	21.59&	24.34 \\ \bottomrule
\end{tabular}
    \begin{tablenotes}\scriptsize
      \item[] Note: 'w/o temp. and ratio' indicates that the temperature parameter ($\tau$, see Eq.~\ref{eq:05}) and the ratio parameter ($\alpha$, see Eq.~\ref{eq:06}) were not used in the POCL framework. Detailed results are listed in Table~\ref{tab:a03}.  
    \end{tablenotes}
\end{threeparttable}
\end{table}

\textbf{Effect of Adaptive Parameter Scheduling.}  
We further analyze how adaptive changes in temperature and SFT ratio impact performance. As illustrated in Figure~\ref{fig:ETS}(b) and Table~\ref{tab:a06} (Appendix~\ref{app:02}), gradually increasing the temperature improves performance across both on-policy and off-policy KD methods. This aligns with the least-to-most prompting principle \cite{libby2008comparison}, where early stages focus on confident predictions, and later stages emphasize soft-label learning from the teacher model.
As illustrated in Figure~\ref{fig:ETS}(c) and Table~\ref{tab:a07} (Appendix~\ref{app:02}), 
for SFT ratio, we find that a gradual decrease benefits off-policy KD by promoting progressive knowledge transfer, while omitting it entirely works best for on-policy KD. This suggests that relying on ground-truth labels can interfere with student learning from teacher-generated outputs in on-policy settings.

\section{Related Work}\label{sec:06} 

\textbf{White-Box KD for LLMs.}  
White-box KD exploits internal teacher model information to enable more effective and interpretable learning compared to black-box methods \cite{yang2024survey}. Recent work has focused on designing loss functions—such as KLD \cite{hinton2015distilling}, RKL \cite{gu2023minillm}, JSD \cite{agarwal2024policy}, and skew KLD variants \cite{ko2024distillm}—to balance mode-seeking and mode-averaging behaviors, reducing issues like mode collapse. Complementary efforts explore data curation strategies, including TGOs and SGOs, to improve training efficiency and align with inference conditions \cite{agarwal2024policy, ko2024distillm, xu2024speculative}. However, identifying optimal combinations of loss functions and data strategies remains an open challenge.

\textbf{CL for KD.}  
While CL has shown benefits in deep learning, its use in KD remains limited. In computer vision, self-paced learning \cite{xiang2020learning} and dynamic temperature-based CL \cite{li2023curriculum} improve robustness by gradually increasing task difficulty. In NLP, Confucius \cite{gao2024confucius} applies multi-stage CL to train LLMs to use complex tools via black-box KD. However, this method is domain-specific and not easily generalizable to white-box KD settings.

\section{Conclusion}\label{sec:07}
In this work, we propose POCL—a plug-and-play curriculum learning framework for white-box knowledge distillation of LLMs, inspired by the “progressive overload” principle in strength training. POCL consists of a Difficulty Measurer that ranks training samples by difficulty and partitions them into ordered subsets, and a Training Scheduler that incrementally introduces subsets from easy to hard while applying loss functions with progressively rising temperatures. This staged exposure stabilizes learning, accelerates convergence, and improves performance across diverse knowledge distillation methods. Extensive experiments demonstrate that POCL consistently enhances the performance and convergence of distilled LLMs, regardless of the underlying knowledge distillation algorithm. Our results highlight the value of structured data ordering in knowledge distillation and provide a practical, generalizable solution for improving white-box distillation of LLMs. Nonetheless, further experimentation across various LLMs and parameter sizes is beneficial for a more comprehensive assessment of the effectiveness and generalizability of the POCL framework (see Appendix~\ref{app:03}).

\bibliographystyle{unsrt}

\appendix
\section{Technical Appendices}
\subsection{Preliminary Formulation of White-box KD}\label{app:00}
We present the background and preliminary formulation of white-box KD for auto-regressive generative student LMs using teacher LLMs. Given a training set composed of source-target sequence pairs $(x, y)$, the student model is trained on two objectives: (1) minimizing the cross-entropy loss between the ground-truth target sequence $y$ and the student model's conditional distribution $q_{\theta}(y|x)$, and (2) minimizing the KD loss that measures the divergence between the teacher model’s token-level distribution $p(y|x)$ and the student’s distribution $q_{\theta}(y|x)$ at each position $i$.

The cross-entropy loss is defined as:

\begin{equation}\label{eq:ce}
L_{ce} = -\sum_{i}^{|y|} \log q_{\theta}(y_i | x, y_{<i}),
\end{equation}

where $q_{\theta}(y_i | x, y_{<i})$ denotes the probability of the student model assigning to token $y_i$ conditioned on the input context $x$ and previous tokens $y_{<i}$. This probability is computed via the softmax function over the student model's logits:

\begin{equation}
q_{\theta}(y_i | x, y_{<i}) = \frac{\exp(z^s_{y_i})}{\sum_{j \in V} \exp(z^s_j)},
\end{equation}

with $z^s_{y_i}$ being the logit corresponding to token $y_i$, and $V$ representing the vocabulary.

The KD loss is formulated as:

\begin{equation}\label{eq:kd}
L_{kd} = -\tau^2 \cdot \sum_{i} D\left(p(y_i | x, y_{<i}; \tau) \parallel q_{\theta}(y_i | x, y_{<i}; \tau)\right),
\end{equation}

where $D(\cdot\parallel\cdot)$ is a divergence measure such as Kullback–Leibler divergence (KLD) and Jensen–Shannon Divergence (JSD), and $\tau$ is distillation temperature coefficient that controls the sharpness of the output distributions. The temperature-scaled probabilities are given by: 

\begin{equation}\label{eq:tem_q}
q_{\theta}(y_i | x, y_{<i}; \tau) = \frac{\exp(z^s_{y_i}/\tau)}{\sum_{j \in V} \exp(z^s_j/\tau)}
\end{equation} and 

\begin{equation}\label{eq:tem_p}
p(y_i | x, y_{<i}; \tau) = \frac{\exp(z^t_{y_i}/\tau)}{\sum_{j \in V} \exp(z^t_j/\tau)}, 
\end{equation}
where $z^t_j$ represents the logits from the teacher model. To ensure numerical stability and comparability between $L_{ce}$ and $L_{kd}$, the KD loss is scaled by $\tau^2$ \cite{hinton2015distilling}. Combining both objectives, the total loss function for the student model becomes:

\begin{equation}\label{eq:total_loss}
L_s = \alpha \cdot L_{ce} + (1 - \alpha) \cdot L_{kd},
\end{equation}
where $\alpha \in [0, 1]$ balances the influence of the SFT and KD components during training.

Recent studies on white-box KD have primarily focused on improving the choice of divergence function $D(\cdot \parallel \cdot)$ in Eq. (\ref{eq:kd}), including standard KLD \cite{hinton2015distilling}, reverse KLD \cite{gu2023minillm}, Jensen-Shannon divergence (JSD) \cite{agarwal2024policy}, and skew KLD \cite{ko2024distillm}. Additionally, various data curation strategies have been explored, such as using ground-truth outputs \cite{hinton2015distilling}, teacher-generated outputs (TGOs) \cite{kim2016sequence}, and student-generated outputs (SGOs) \cite{lin2020autoregressive, agarwal2024policy}, or combinations thereof \cite{gu2023minillm, ko2024distillm, ko2025distillm}.

\subsection{Experimental Setup}\label{app:01}

This appendix provides detailed information on the base models (Section~\ref{app:0101}), training procedures (Section~\ref{app:0102}), evaluation protocols (Section~\ref{app:0103}), and baseline methods (Section~\ref{app:0104}).

\subsubsection{Base Models Description}\label{app:0101}

We select two widely used large language model (LLM) families—GPT-2 \cite{radford2019language} and OPT \cite{zhang2022opt}—at different scales for our experiments:

\begin{itemize}
    \item \textbf{GPT-2} \cite{radford2019language}: A decoder-only transformer-based language model developed by OpenAI. In our experiments, GPT-2 (0.1B) and GPT-2 (1.5B) are used as student and teacher models, respectively. Pretrained checkpoints are available at \url{https://huggingface.co/openai-community/gpt2}.
    
    \item \textbf{OPT} \cite{zhang2022opt}: A family of open-source decoder-only LLMs developed by Meta AI, ranging from 125 million to 175 billion parameters. We use OPT (350m) and OPT (2.7B) as student and teacher models, respectively. Checkpoints are publicly accessible at \url{https://huggingface.co/facebook/opt-350m}.
\end{itemize}

These models were chosen for their representativeness and open availability, enabling reproducibility and comparison.

\subsubsection{Training Details}\label{app:0102}

All experiments are conducted using four A800 80GB GPUs and an Intel(R) Xeon(R) Platinum 8350C CPU. We follow standard practices for instruction-following tasks, using the \texttt{databricks-dolly-15K} dataset \cite{DatabricksBlog2023DollyV2}, which contains 15,000 human-written instruction-response pairs. Samples exceeding the maximum context length are removed to ensure compatibility with model constraints. The dataset is randomly split into 12.5K training, 1K validation, and 0.5K test samples.

Hyperparameters—including learning rates (\{5e-4, 1e-4, 5e-5\}) and batch sizes (\{8, 16\})—are selected based on validation performance. For a fair comparison, we maintain an equal total training steps between the baseline models and our proposed framework. In particular, whereas the baseline models are trained for 20 epochs, our framework are trained for 8 epochs under the same experimental configuration. Final checkpoints are selected using Rouge-L scores on the validation set, following evidence that Rouge-L correlates well with human judgments \cite{agarwal2024policy}.

For on-policy knowledge distillation (KD), we use a mixture of 50\% student-generated outputs (SGOs) and ground-truth responses, as recommended by \cite{agarwal2024policy}. 

In the POCL framework:
\begin{itemize}
    \item Training samples are ranked by difficulty using Eq.~(\ref{eq:04}) and divided into $n = 4$ subsets.
    \item Trained for 40\% epochs of baselines to maintain equal total training steps of baselines. 
    \item Each stage uses the final checkpoint of the previous stage as its initialization.
    \item Adaptive temperature parameters are initialized with $\tau_{0} = 1$ and increased to $\tau_{n} = 2$.
    \item For off-policy KD, the SFT ratio decreases linearly from $\alpha_{0} = 0.3$ to $\alpha_{n} = 0$.
    \item For on-policy KD, no adaptive SFT ratio is applied ($\alpha_{0} = \alpha_{n} = 0$).
\end{itemize}

These settings reflect our empirical findings and aim to balance convergence speed and performance stability. The time required to run an experiment varies depending on the LLMs and base KD methods. Specifically, it costs around ten minutes to run a GPT2(0.1B) KLD-based POCL once.

\subsubsection{Evaluation Details}\label{app:0103}
\begin{figure}[!ht]
   \begin{center}
     \includegraphics[scale = 0.1]{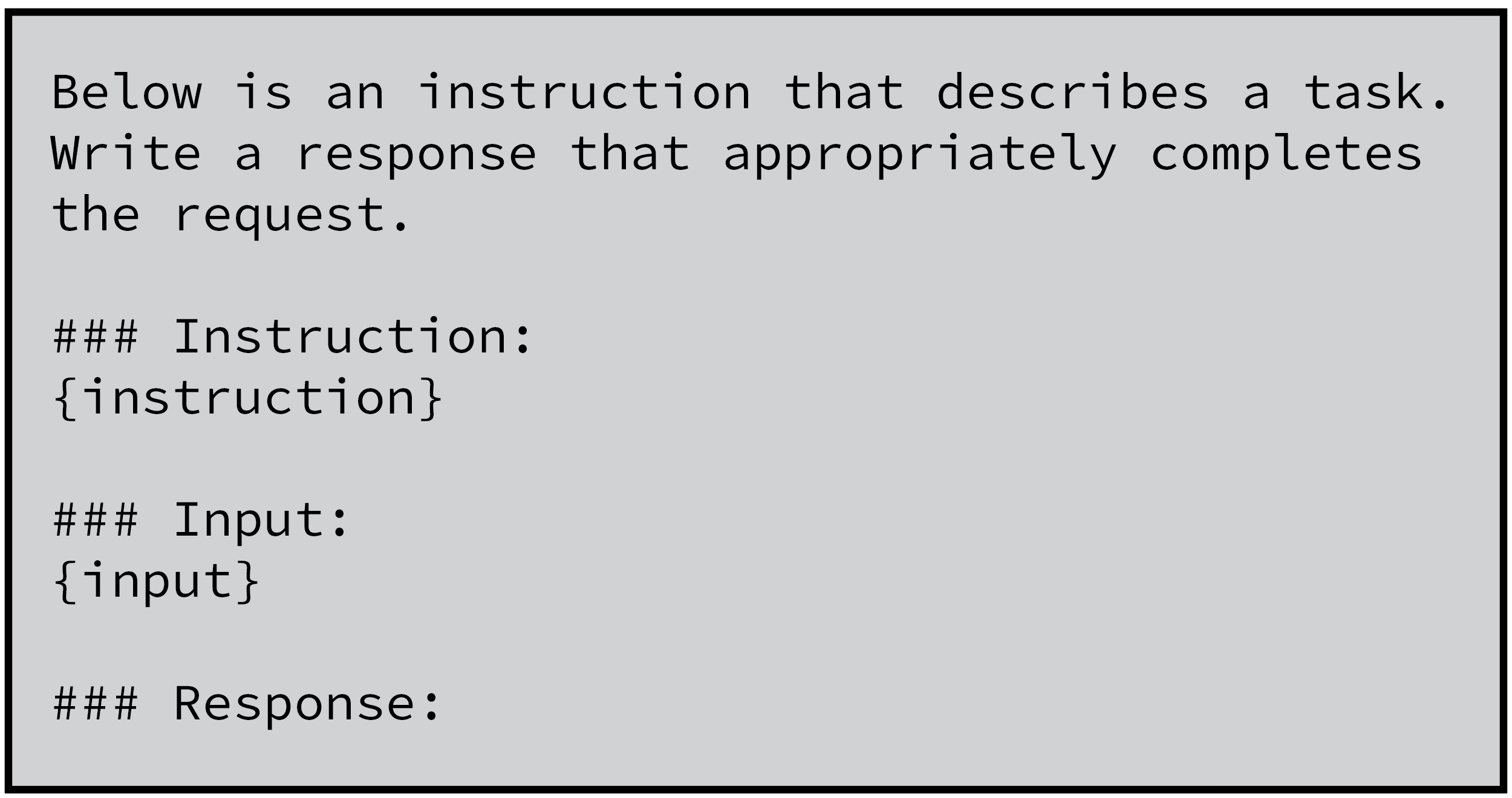}
   \caption{\small Prompt template used for training and evaluation in instruction-following experiments, adapted from \cite{gu2023minillm}}\label{fig:prompt}
   \end{center}
\end{figure}

All model evaluations are conducted on a single A800 80GB GPU, following the setup described in \cite{gu2023minillm}. We use a fixed prompt template during evaluation, as illustrated in Fig.~\ref{fig:prompt}, to ensure consistency across models.

During inference, responses are generated using a temperature of 1.0 and a maximum sequence length of 512 tokens. To ensure robustness, we generate five responses per prompt using different random seeds (\{10, 20, 30, 40, 50\}) and report the average performance across these samples.

Below, we provide descriptions of the datasets used for training and evaluation:

\begin{itemize}
    \item \texttt{databricks-dolly-15K} \cite{DatabricksBlog2023DollyV2}: An open-source dataset containing 15,000 human-written instruction-response pairs across diverse categories such as brainstorming, classification, closed QA, generation, information extraction, open QA, and summarization \cite{ouyang2022training}.

    \item \texttt{self-instruct-eval} \cite{wang2022self}: A framework that leverages model-generated outputs to synthesize instructional data. It includes 52,000 instructions and 82,000 input-output pairs for training, along with 252 expert-written tasks and an additional 50,000 public examples for evaluation.

    \item \texttt{vicuna-eval} \cite{chiang2023vicuna}: A benchmark consisting of 80 challenging questions originally used to evaluate Vicuna, designed to test complex reasoning and instruction-following capabilities.

    \item \texttt{supernatural-instructions} \cite{wang2022benchmarking}: A comprehensive benchmark comprising 1,616 distinct NLP tasks with expert-written instructions, spanning 76 task types. The test set contains approximately 9,000 samples drawn from 119 tasks.

    \item \texttt{unnatural-instructions-core} \cite{honovich2022unnatural}: A large-scale dataset of 240,000 machine-generated instructions, demonstrating that high-quality synthetic data can rival human-written data for training language models. The core subset consists of 66,000 samples.
\end{itemize}

\subsubsection{Baseline Description}\label{app:0104}
This section provides formal definitions of the knowledge distillation (KD) loss functions used in our experiments. As described in Eq.~(\ref{eq:kd}), the KD loss measures the divergence between the teacher model distribution $p$ and the student model distribution $q_{\theta}$ using a distance function $D(\cdot||\cdot)$. Below are the specific formulations considered in this study.

The Kullback–Leibler Divergence (KLD) \cite{hinton2015distilling} is defined as:

\begin{equation}\small\label{eq.07}
    D_{KLD}(p \parallel q_{\theta}) = \mathbb{E}_{y \sim p} \left[\log \frac{p(y)}{q_{\theta}(y)} \right],
\end{equation}

where $y$ is sampled from the teacher distribution $p$.

The Reverse KLD (RKL) is given by:

\begin{equation}\small\label{eq.08}
    D_{RKL}(q_{\theta} \parallel p) = \mathbb{E}_{y \sim p} \left[\log \frac{q_{\theta}(y)}{p(y)} \right].
\end{equation}

The Jensen–Shannon Divergence (JSD) \cite{agarwal2024policy} is formulated as:

\begin{equation}\small\label{eq.09}
    D_{JSD} = \frac{1}{2} \mathbb{E}_{y \sim p} \left[\log \frac{p(y)}{m(y)} \right] + \frac{1}{2} \mathbb{E}_{y \sim p} \left[\log \frac{q_{\theta}(y)}{m(y)} \right],
\end{equation}

where $m(\cdot) = \frac{1}{2}p(\cdot) + \frac{1}{2}q_{\theta}(\cdot)$ is the average of the two distributions.

The Total Variation Distance (TVD) \cite{wen2023f} is defined as:

\begin{equation}\small\label{eq.10}
    D_{TVD} = \frac{1}{2}\sum_{y}|q_{\theta}(y) - p(y)|.
\end{equation}

The Forward Skew KLD (SKL) \cite{ko2024distillm} is expressed as:

\begin{equation}\small\label{eq.11}
    D_{SKL}(p \parallel q_{\theta}) = D_{KLD}\left(p \parallel \beta \cdot p + (1-\beta) \cdot q_{\theta} \right),
\end{equation}

and the Reverse Skew KLD (SRKL) \cite{ko2024distillm} is:

\begin{equation}\small\label{eq.12}
    D_{SRKL}(q_{\theta} \parallel p) = D_{KLD}\left(q_{\theta} \parallel (1 - \beta) \cdot p + \beta \cdot q_{\theta} \right),
\end{equation}

where $\beta \in [0,1]$ controls the mixing ratio between the teacher and student distributions.

These loss functions represent commonly used strategies for aligning student and teacher distributions in white-box KD.

\subsection{Additional Results}\label{app:02}
In this section, we present additional experimental results to further validate the effectiveness of our proposed POCL framework and its key components.

In table \ref{tab:a01}, We present evaluation results showing ROUGE-L scores across five random seeds for opt-350 as the student, to show the effectiveness and generalizability of POCL framework on KD across different LLMs.  

\begin{table}\centering\scriptsize
\caption{Evaluation results showing ROUGE-L scores across five random seeds for models trained on the test set, based on the OPT family.}\label{tab:a01}
\begin{threeparttable}
\begin{tabular}{ccl|cccccc} \toprule
Model & \# Params & KD Methods & \textbf{DollyEval} & \textbf{SelfInst} & \textbf{VicunaEval} & \textbf{S-NI} & \textbf{UnNI} & \textbf{Avg.} \\ \midrule
                        &                     2.7B & Teacher                      &26.79&15.22&17.05&23.79&28.07&22.18\\ \cmidrule(r){2-9} 
                        &                          & SFT                          &21.48&11.06&14.21&18.33&21.69&17.35\\
                        &                          & SeqKD                        &23.01&10.19&14.31&18.93&21.58&17.60\\ 
                        &                          & GKD                          &24.95&13.53&17.46&21.77&25.66&20.67\\
                        &                          & ~~+POCL                      &25.87&14.11&17.65&23.08&27.56&21.65 (0.98 $\uparrow$)\\
                        &                          & KLD                          &23.09&10.03&14.17&19.11&22.68&17.82\\
                        &                          & ~~+POCL                      &25.45&11.59&16.45&21.42&24.93&19.97 (2.15 $\uparrow$)\\
                        &                          & RKL                          &23.88&12.01&16.49&23.92&22.78&19.82\\
                        &                          & ~~+POCL                      &25.21&12.45&16.53&23.57&23.77&20.31 (0.48 $\uparrow$)\\        
                        &                          & JSD                          &23.97&12.97&17.03&23.51&27.08&20.91\\
                        &                          & ~~+POCL                      &25.51&12.57&\textbf{17.86}&23.30&29.22&21.69 (0.77 $\uparrow$)\\
                        &                          & TVD                          &24.82&13.63&15.87&27.66&30.31&22.45\\
                        &                          & ~~+POCL                      &24.81&\textbf{14.82}&16.36&\textbf{28.24}&\textbf{31.12}&\textbf{23.06} (0.61 $\uparrow$)\\
                        &                          & SKL                          &25.02&12.66&16.29&23.46&28.67&21.22\\
                        &                          & ~~+POCL                      &\textbf{26.13}&12.84&17.21&25.62&30.02&22.36 (1.13 $\uparrow$)\\
                        &                          & SRKL                         &24.88&14.66&16.56&23.95&28.12&21.63\\
\multirow{-16}{*}{OPT} & \multirow{-15}{*}{0.3B}   & ~~+POCL                      &25.21&14.17&17.05&25.84&29.97&22.45 (0.81 $\uparrow$)\\ \bottomrule
\end{tabular}
    \begin{tablenotes}\scriptsize
      \item[] Note: GKD \cite{hinton2015distilling} uses JSD as the KD loss and employs an on-policy data curation strategy, whereas JSD refers to the use of JSD as the KD loss with an off-policy data curation approach. The best scores among student models are \textbf{bold}. RKL, TVD, SKL, and SRKL denote reverse KLD, total variation distance, forward and reverse skew KLD, respectively. \textbf{Avg.} denotes the average ROUGE-L score across five evaluation datasets.
    \end{tablenotes}
\end{threeparttable}
\end{table}

Table~\ref{tab:a04} shows the performance comparison of KLD and KLD+POCL using different ranking strategies—fusion, Rouge-L only, and cross-entropy only—on out-of-domain evaluation datasets. These results complement our earlier analysis by demonstrating the generalization benefits of the fusion-based ranking across unseen domains.

\begin{table}\centering\small
\begin{threeparttable}
\caption{Performance comparison of KLD and KLD+POCL using different ranking approaches on the other evaluation datasets.}\label{tab:a04}
\begin{tabular}{c|l|cc} \toprule
Evaluation  & Model &  GPT-2 & OPT \\ 
Dataset & \# Params (T/S) & 1.5B/0.1B & 2.7B/0.3B  \\ \midrule
\multirow{4}{*}{SelfInst} & KLD  & 10.33&	10.03 \\
&~~+POCL (R-L)                   & 11.10&	10.83 \\
&~~+POCL (Loss)                  & 11.79&	11.03 \\
&~~+POCL (Fusion)                & 11.56&	11.59 \\ \midrule
\multirow{4}{*}{VicunaEval} & KLD& 14.96&	14.17 \\
&~~+POCL (R-L)                   & 16.24&	15.66 \\
&~~+POCL (Loss)                  & 15.55&	16.16 \\
&~~+POCL (Fusion)                & 16.13&	16.45 \\ \midrule
\multirow{4}{*}{S-NI} & KLD      & 19.70&	19.11 \\
&~~+POCL (R-L)                   & 21.06&	21.65 \\
&~~+POCL (Loss)                  & 21.93&	20.39 \\
&~~+POCL (Fusion)                & 21.59&	21.42 \\ \midrule
\multirow{4}{*}{UnNI} & KLD      & 22.01&	22.68 \\
&~~+POCL (R-L)                   & 24.60&	23.78 \\
&~~+POCL (Loss)                  & 24.72&	24.57  \\
&~~+POCL (Fusion)                & 24.34&	24.93  \\ \bottomrule
\end{tabular}
    \begin{tablenotes}\scriptsize
      \item[] Note: 'Params (T/S)' denotes the number of parameters in the teacher and student models. 'R-L' indicates that training samples are sorted by ROUGE-L score \cite{lin2004rouge}, 'Loss' by cross-entropy loss, and 'Fusion' by the Fusion rank (see Eq. \ref{eq:04}).
    \end{tablenotes}
\end{threeparttable}
\end{table}

Table~\ref{tab:a03} presents an ablation study on adaptive parameter scheduling within POCL, comparing GKD+POCL and JSD+POCL with and without distillation temperature and SFT ratio adjustments. This highlights the contribution of dynamic parameter control to performance improvements under both off-policy and on-policy KD settings.

\begin{table}\centering\small
\begin{threeparttable}
\caption{Performance comparison of GKD+POCL \& JSD+POCL with and without adaptive parameters on the five evaluation datasets.}\label{tab:a03}
\begin{tabular}{l|ccccc} \toprule
KD Methods  & \textbf{DollyEval} & \textbf{SelfInst} & \textbf{VicunaEval} & \textbf{S-NI} & \textbf{UnNI} \\ \midrule
 GKD                            &24.67& 	11.48& 15.66&  23.80&  25.26  \\
~~+POCL (w/o temp.)             &24.05&  11.09&	15.27&	23.21&	26.01  \\
~~+POCL                         &26.60&  12.62&	16.70&	27.02&	29.61  \\ \midrule
 JSD                            &23.79& 	11.38& 15.87&  22.84&  23.06  \\
~~+POCL (w/o temp. \& ratio)    &22.73&  10.94&	15.21&	21.99&	22.65  \\
~~+POCL (w/o temp.)             &23.63&  11.47&	15.86&	23.08&	22.93  \\
~~+POCL (w/o ratio)             &25.87&  11.72&	16.68&	26.41&	24.78  \\
~~+POCL                         &25.97&  11.74&	16.77&	26.61&	25.21  \\ \bottomrule
\end{tabular}
    \begin{tablenotes}\scriptsize
      \item[] Note: GKD \cite{hinton2015distilling} uses JSD as the KD loss and employs an on-policy data curation strategy, whereas JSD refers to the use of JSD as the KD loss with an off-policy data curation approach. 
    \end{tablenotes}
\end{threeparttable}
\end{table}

The impact of different training sample ordering strategies is summarized in Table~\ref{tab:a05}, showing how easy-to-hard versus hard-to-easy scheduling affects distillation outcomes across multiple KD methods and model architectures.

\begin{table}\centering\small
\begin{threeparttable}
\caption{Results showing the impact of different training sample ranking strategies on the performance of the distilled student model across various KD methods.}\label{tab:a05}
\begin{tabular}{ccl|ccccc} \toprule
Teacher & Student & KD Methods  & \textbf{DollyEval} & \textbf{SelfInst} & \textbf{VicunaEval} & \textbf{S-NI} & \textbf{UnNI} \\ \midrule
      &        & GKD            & 24.67&	 11.48&	15.66&	23.80&	25.26   \\
      &        & ~~+POCL (E2H)  & 26.60&	 12.62&	16.70&	27.02&	29.61   \\
GPT2  & GPT2   & ~~+POCL (H2E)  & 24.73&   12.56& 16.37&  24.35&  25.28   \\ \cmidrule(r){3-8}
(1.5B)& (0.1B) & KLD            & 23.49&   10.33& 14.96&  19.70&  22.01   \\
      &        & ~~+POCL (E2H)  & 24.87&	 11.56&	16.13&	21.59&	24.34   \\
      &        & ~~+POCL (H2E)  & 24.03&	 11.51&	16.04&	20.63&	23.36   \\ \midrule
      &        & GKD            & 24.95&	 13.53&	17.46&	21.77&	25.66   \\
      &        & ~~+POCL (E2H)  & 25.87&	 14.11&	17.65&	23.08&	27.56   \\
OPT   & OPT    & ~~+POCL (H2E)  & 25.33&   13.75& 17.47&  22.86&  26.75   \\ \cmidrule(r){3-8}
(2.7B)& (0.3B) & KLD            & 23.09&   10.03& 14.17&  19.11&  22.68   \\
      &        & ~~+POCL (E2H)  & 25.45&	 11.59&	16.45&	21.42&	24.93   \\
      &        & ~~+POCL (H2E)  & 25.35&	 10.19&	16.61&	19.41&	24.54   \\ \bottomrule
\end{tabular}
    \begin{tablenotes}\scriptsize
      \item[] Note: 'E2H' and 'H2E' denote that training samples are sorted from easy to hard and hard to easy, respectively, to construct training subsets in the POCL framework for knowledge distillation.
    \end{tablenotes}
\end{threeparttable}
\end{table}

Tables~\ref{tab:a06} and \ref{tab:a07} evaluate the effects of varying temperature scaling and SFT ratio schedules, respectively. These results illustrate how adaptive control of individual components within POCL consistently enhances knowledge transfer across diverse student models and distillation objectives.

\begin{table}\centering\small
\begin{threeparttable}
\caption{Results showing the impact of different temperature change strategies on the performance of the distilled student model across various KD methods.}\label{tab:a06}
\begin{tabular}{ccl|ccccc} \toprule
Teacher & Student & KD Methods  & \textbf{DollyEval} & \textbf{SelfInst} & \textbf{VicunaEval} & \textbf{S-NI} & \textbf{UnNI} \\ \midrule
      &        & GKD                  & 24.67&	11.48&	15.66&	23.80&	25.26    \\
      &        & ~~+POCL (temp. 2-1)  & 23.93&	10.95&	15.18&	23.06&	25.33    \\
GPT2  & GPT2   & ~~+POCL (temp. 1-2)  & 26.60&	  12.62&  16.70&  27.02&  29.61    \\ \cmidrule(r){3-8}
(1.5B)& (0.1B) & KLD                  & 23.49&	  10.33&  14.96&  19.70&  22.01    \\
      &        & ~~+POCL (temp. 2-1)  & 21.92&	 9.23&	13.87&	18.97&	21.62    \\
      &        & ~~+POCL (temp. 1-2)  & 24.87&	11.56&	16.13&	21.59&	24.34    \\ \midrule
      &        & GKD                  & 24.95&	13.53&	17.46&	21.77&	25.66    \\
      &        & ~~+POCL (temp. 2-1)  & 24.05&	13.76&	17.47&	21.73&	25.65    \\
OPT   & OPT    & ~~+POCL (temp. 1-2)  & 25.87&	  14.11&  17.65&  23.08&  27.56    \\ \cmidrule(r){3-8}
(2.7B)& (0.3B) & KLD                  & 23.09&	  10.03&  14.17&  19.11&  22.68    \\
      &        & ~~+POCL (temp. 2-1)  & 23.27&	 9.61&	13.79&	19.47&	22.12    \\
      &        & ~~+POCL (temp. 1-2)  & 25.45&	11.59&	16.45&	21.42&	24.93    \\ \bottomrule
\end{tabular}
    \begin{tablenotes}\scriptsize
      \item[] Note: 'temp. 2-1' and 'temp. 1-2' indicate that the initial ($\tau_{0}$) and final ($\tau_{n}$) temperatures are set to 2 and 1, and 1 and 2, respectively; the temperature values are computed according to Eq. \ref{eq:05}.
    \end{tablenotes}
\end{threeparttable}
\end{table}

\begin{table}\centering\small
\begin{threeparttable}
\caption{Results showing the impact of different STF ratio change strategies on the performance of the distilled student model across various KD methods.}\label{tab:a07}
\begin{tabular}{ccl|ccccc} \toprule
Teacher & Student & KD Methods  & \textbf{DollyEval} & \textbf{SelfInst} & \textbf{VicunaEval} & \textbf{S-NI} & \textbf{UnNI} \\ \midrule
      &        & GKD                    & 24.67&	11.48&	15.66&	23.80&	25.26    \\
      &        & ~~+POCL (ratio 0-0.3)  & 25.26&	11.55&	16.19&	26.95&	27.79    \\
GPT2  & GPT2   & ~~+POCL (ratio 0.3-0)  & 26.60&  12.62&  16.70&  27.02&  29.61    \\ \cmidrule(r){3-8}
(1.5B)& (0.1B) & KLD                    & 23.49&  10.33&  14.96&  19.70&  22.01    \\
      &        & ~~+POCL (ratio 0-0.3)  & 24.20&	11.75&	15.36&	21.53&	22.20    \\
      &        & ~~+POCL (ratio 0.3-0)  & 24.87&	11.56&	16.13&	21.59&	24.34    \\ \midrule
      &        & GKD                    & 24.95&	13.53&	17.46&	21.77&	25.66    \\
      &        & ~~+POCL (ratio 0-0.3)  & 25.75&	14.11&	17.69&	22.84&	26.82    \\
OPT   & OPT    & ~~+POCL (ratio 0.3-0)  & 25.87&  14.11&  17.65&  23.08&  27.56    \\ \cmidrule(r){3-8}
(2.7B)& (0.3B) & KLD                    & 23.09&  10.03&  14.17&  19.11&  22.68    \\
      &        & ~~+POCL (ratio 0-0.3)  & 24.63&	10.56&	15.75&	19.15&	23.01    \\
      &        & ~~+POCL (ratio 0.3-0)  & 25.45&	11.59&	16.45&	21.42&	24.93    \\ \bottomrule
\end{tabular}
    \begin{tablenotes}\scriptsize
      \item[] Note: 'ratio 0-0.3' and 'ratio 0.3-0' indicate that the initial ($\alpha_{0}$) and final ($\alpha_{n}$) SFT ratio are set to 0 and 0.3, and 0.3 and 0, respectively; the ratio values are computed according to Eq. \ref{eq:06}.
    \end{tablenotes}
\end{threeparttable}
\end{table}

\subsection{Limitations }\label{app:03}
Due to computational constraints, our experiments were limited to two model families—GPT-2 and OPT—to evaluate the effectiveness of the POCL framework in boosting KD performance. Future work should investigate additional model families, such as Qwen3 \cite{qwen3} and LLaMA 3 \cite{meta2024llama32}, to assess the generalizability of POCL across diverse architectures. Furthermore, within each model family, we considered only a single parameter configuration. Exploring varying model sizes could provide deeper insights into how POCL interacts with model capacity.
For example, student LMs with larger parameter sizes may inherently possess sufficient capacity to understand the training set, leading to a narrow distribution of perceived difficulty levels among training samples when measured and ranked by the model. This limited diversity in difficulty estimation could, in turn, affect the effectiveness of the POCL framework. We plan to address these limitations in future work by conducting more comprehensive experiments across model architectures and scales.

\end{document}